\documentclass[letterpaper, 10 pt, conference]{ieeeconf}
\IEEEoverridecommandlockouts
\usepackage{graphicx} 
\usepackage{amsmath} 
\usepackage{amssymb}  
\usepackage{booktabs}
\usepackage{subcaption}
\usepackage{mleftright}
\usepackage{floatrow}

\newfloatcommand{capbtabbox}{table}[][\FBwidth]

\title{\LARGE \bf
First-Person Perceptual Guidance Behavior Decomposition using Active Constraint Classification
}

\author{Andrew Feit$^{1}$ and B{\'e}r{\'e}nice Mettler $^{2}$
\thanks{*This work was supported by a grant from the Office of Naval Research}
\thanks{$^{1}$Andrew Feit is a graduate student with the Department of Aerospace Engineering, University of Minnesota, Minneapolis, MN 55401, USA {\tt\small feit0003@umn.edu}}%
\thanks{$^{2}$B{\'e}r{\'e}nice Mettler is an adjunct associate professor with the Department of Aerospace Engineering, University of Minnesota, Minneapolis, MN 55401, USA, she is also a faculty reearcher at ICSI, Berkeley, CA. {\tt\small mettler@umn.edu}}%
}

\begin{document}

\maketitle
\thispagestyle{empty}
\pagestyle{empty}

\begin{abstract}
Humans exhibit a wide range of adaptive and robust dynamic motion behavior that is yet unmatched by autonomous control systems. These capabilities are essential for real-time behavior generation in cluttered environments. Recent work suggests that human capabilities rely on task structure learning and embedded or ecological cognition in the form of perceptual guidance. This paper describes the experimental investigation of the functional elements of human motion guidance, focusing on the control and perceptual mechanisms. The motion, control, and perceptual data from first-person guidance experiments is decomposed into elemental segments based on invariants. These elements are then analyzed to determine their functional characteristics. The resulting model explains the structure of the agent-environment interaction and provides lawful descriptions of specific perceptual guidance and control mechanisms.
\end{abstract}


\section{Introduction}

Trained humans are capable of a wide range of high-performance, robust, and adaptive motion behavior that exceeds the capabilities of current autonomous systems. Recent advancements in deep learning demonstrate impressive performance by taking advantage of task structure \cite{bojarski2016end}, but are not yet able to explain the principles underlying the control capabilities. An explicit description of the structural elements that are responsible for this behavior is critical for intelligent machines to be able to interact with human operators, and to be able to verify the performance of a system over a task domain.

The present works seeks to identify specific functional elements in a simulated driving task based on equivalence classes \cite{mettler2015systems} in the human-environment system dynamics. The decomposition is formulated using a hierarchical model for human dynamic motion behavior \cite{mettler2013hierarchical} that divides human motion into fundamental levels of behavior: planning, guidance, and tracking. Experimental data is first decomposed into segments belonging to equivalence classes that correspond to distinct patterns in guidance behaviors. These patterns are referred to as interaction patterns (IP), and have previously been proposed to serve as a unit of organization of behavior. Because IP serve essential function roles in human behavior, they are used here as the unit of analysis \cite{kong2013modeling}.

Previous work modeled human interaction patterns as a control policy \cite{mettler2013mapping, kong2013modeling, feit2016extraction} that specifies the optimal action over a spatial task domain. A spatial value function (SVF) representation has limitations however: first, an agent using an SVF is only capable of choosing actions within the represented function domain. Second, large spatial domains associated with tasks such as driving or piloting would require impractical amounts of memory and a huge amount of prior experience to adequately learn an SVF. In contrast, humans generate motion behavior in large domains and with a small number of example trials suggesting that they extrapolate using functional principles learned from prior experience.

In this work, motion behavior is decomposed into segments that are functionally equivalent, i.e. share similar perception-action signal relationships, and that have similar sets of active constraints. Prior work modeled planning behavior using subgoals that are defined as spatial constraint transition points \cite{feit2017subgoal, feit2015human}. This concept is extended to suggest that transitions between dynamic perceptual guidance control modes occur when the set of active dynamic constraints changes. In addition, the concept of perceptual guidance \cite{lee1976theory} emphasizes the importance of the perception-action information exchange in human behavior \cite{warren2006dynamics, tishby2011information}. This information-based approach suggests that control modes can be identified by unique sets of statistical dependencies between perceptual and action variables.

\section{Related Work}

This section reviews prior work that investigates the structure of human motion guidance behavior and approaches to behavior analysis and segmentation.

\subsection{Constrained Optimal Control}

Prior work showed how a constrained optimal control task can be simplified by segmenting it at constraint transition points \cite{feit2015human, feit2017subgoal}. This approach of dividing a continuous control task into a sequence of elements is similar to the maneuver automaton (MA) model \cite{gavrilets2004human, frazzoli2005maneuver, mettler2002rotorcraft} for motion behavior. A library of maneuver elements simplifies planning by reducing an infinite-dimension trajectory optimization to a finite-dimension planning task that can be solved using dynamic programming. The present work extends the concept of MA to consider guidance elements that incorporate agent-environment interactions including the key functional elements of the perception-action system.

\subsection{Human Motion Guidance}

Prior work in cognitive science and control engineering have investigated the cognitive processes supporting the capabilities needed to perform complex spatial tasks.

\subsubsection{Perceptual Guidance}

Lee's work on perceptual guidance \cite{lee1976theory} extends earlier ideas by Gibson on perception during locomotion \cite{gibson1958visually}, showing that the instantaneous time to gap closure (Tau) is related to observed biological motion profiles and is readily available in visual cues. Tau provides a simple method to generate actions that satisfy task constraints \cite{simon1972theories}. Gibson extended this concept with the idea of ecological cognition \cite{gibson1979ecological}, suggesting that organisms use their physical situation in the environment to generate motion. Gibson \cite{gibson1977theory} and Warren \cite{warren1984perceiving} investigate affordances, suggesting that humans perceive their environment in terms of the results of possible actions. The concept of embodied cognition \cite{wilson2013embodied} extends these ideas, saying that the ecological nature of the agent-environment interaction replaces the need for internal cognitive representations of a task.

\subsubsection{Interaction Patterns}

The concept of interaction patterns (IPs) was introduced by Mettler and Kong~\cite{kong2013modeling,mettler2013hierarchical} for the analysis and modeling of agile spatial guidance behavior. The IPs capture the structure of dynamic interactions and how humans use these properties, in combination with more general characteristics such as invariants and equivalence relations, to help mitigate complexity typically associated with spatial planning and control.

The central role of IPs as units of organization was subsequently used to formulate a hierarchical model that delineates the primary functions and how they are integrated, and details the key sensory and control quantities at each hierarchical control level~\cite{mettler2015systems}. The model also resolves the typical gap between discrete planning and continuous guidance and control.

Other applications of this framework over the past five years further validate that emergent patterns in agent-environment behavior are exploited by human operator \cite{li2015towards,verma2016investigating,tseng2016human}. These elements represent functional units that can be exploited to provide detailed understanding of the underlying control, perceptual and cognitive functions and their organization. 

\subsection{Human Perception}

To investigate how guidance behavior elements relate to both optimal control and to perceptual capabilities, it is necessary to understand the sensory functions required for motion guidance. An important aspect of spatial behavior is the perception-action dynamics. Existing models typically focus on a primary control loop (see Warren~\cite{warren2006dynamics}), however, as highlighted in the functional model~\cite{mettler2013hierarchical}, perception is also a hierarchical process which can be conceived as a multiple-loop system, with specific attentional elements and perceptual functions at each level of the control hierarchy, as described in the hierarchical functional model~\cite{mettler2015systems}. 

This model was investigated using a remote-control helicopter guidance task~\cite{andersh2014modeling}. Experimental data was used to identify perceptual functions such tracking and estimating the current position of the aircraft (via smooth pursuits), and measuring gaps between the current position and a goal or obstacle (via saccades). Additional aspects of perception and human gaze behavior were investigated using simulated navigation tasks in an unknown environment~\cite{verma2015investigation}. 

\subsection{Information Model}

More recently, researchers have taken an information-theoretic perspective on perception and action. Tishby and Polani hypothesize that optimal behavior maximizes both utility and agent-environment information channel capacity and modeled perception and action as a dynamic Bayesian network \cite{tishby2011information}. Polani also investigated the information processing requirements of biological systems during motion control tasks \cite{polani2011informational} and concluded that embodied guidance strategies reduce information processing requirements. Additional research further validates that information processing cost, in addition to utility, may shape human and animal control strategies \cite{drugowitsch2012cost}.

\subsection{Structure Learning}

One important aspect of learning is the ability to recognize subtasks across a domain for which similar policies apply. Braun et al. \cite{braun2010structure} describes structure learning as learning to learn, which can be described as extracting invariants across task episodes that allow knowledge to be shared. Interaction patterns described in~\cite{kong2013modeling} describe a class of invariant structural characteristics of behavior. In contrast to equivalence concepts used in~\cite{kong2013modeling} to identify structure, Van Dijk et al. \cite{dijk2011grounding} describe an information-based method for identifying subgoals in a guidance task modeled as a Markov decision process (MDP). The idea is that control actions depend on relevant goal information (RGI). When the agent reaches a subgoal, the relevant goal information rapidly switches, which is reflected in control action.

\subsection{Data Segmentation and Clustering}

Extracting structural elements in behavior involves identifying and clustering data segments belonging to similar state subspace manifolds. One approach to this problem is piecewise affine clustering (PWA) \cite{ferrari2003clustering}, which has been used for surgical skill analysis \cite{li2015towards}. The PWA process identifies a linear model describing local windows of data points, then identifies a small number of clusters that represent the data. Other work on surgical skill analysis improves the state estimate by assuming a regular transition sequence across the data and excluding behavioral loops \cite{krishnan2018transition}. 

One approach to identify signal relationships within a sample window is sparse graphical modeling \cite{sun2009mining}. Sparse inverse covariance estimate (SICE) uses an optimization procedure to suppresses covariance matrix elements and promote sparsity. For example, Tseng and Mettler~\cite{tseng2016human} applied this SICE to extract sensory-motor patterns from data obtained in a tele-robotic search experiment.

\section{Experimental Setup}

\begin{figure}[tbph]
\centering
\begin{subfigure}{0.62\textwidth}
\centering
\includegraphics[height=3cm]{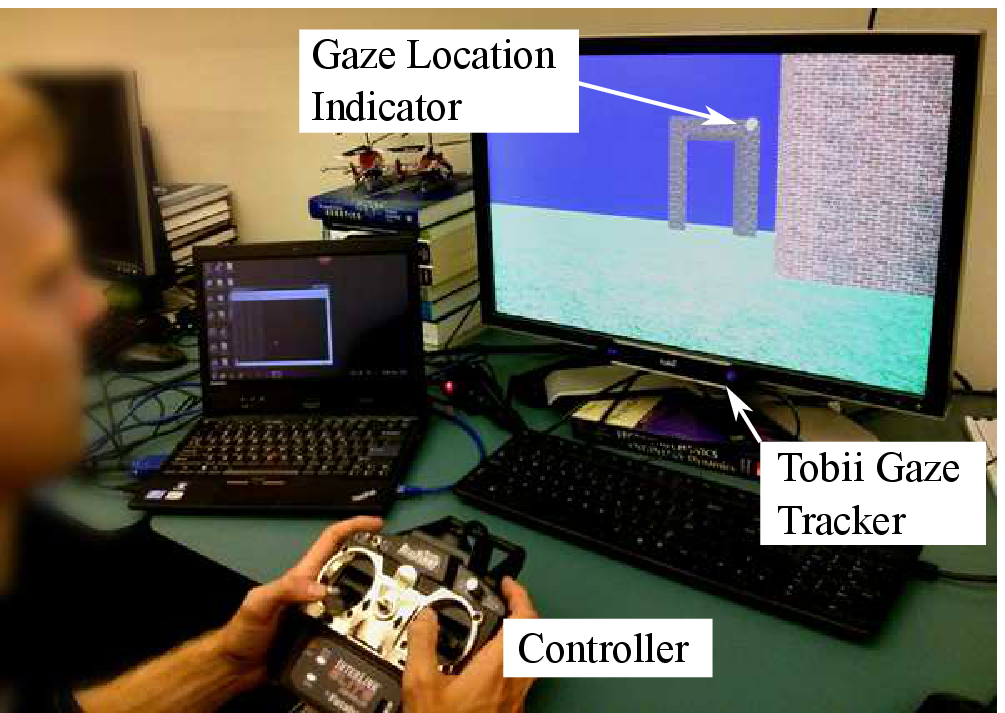}
\caption{First-person experimental setup.}
\label{fig:simulationsetupblurlabelscolor}
\end{subfigure}
\hfil
\begin{subfigure}{0.35\textwidth}
\centering
\vspace{1mm}
\includegraphics[height=3.2cm]{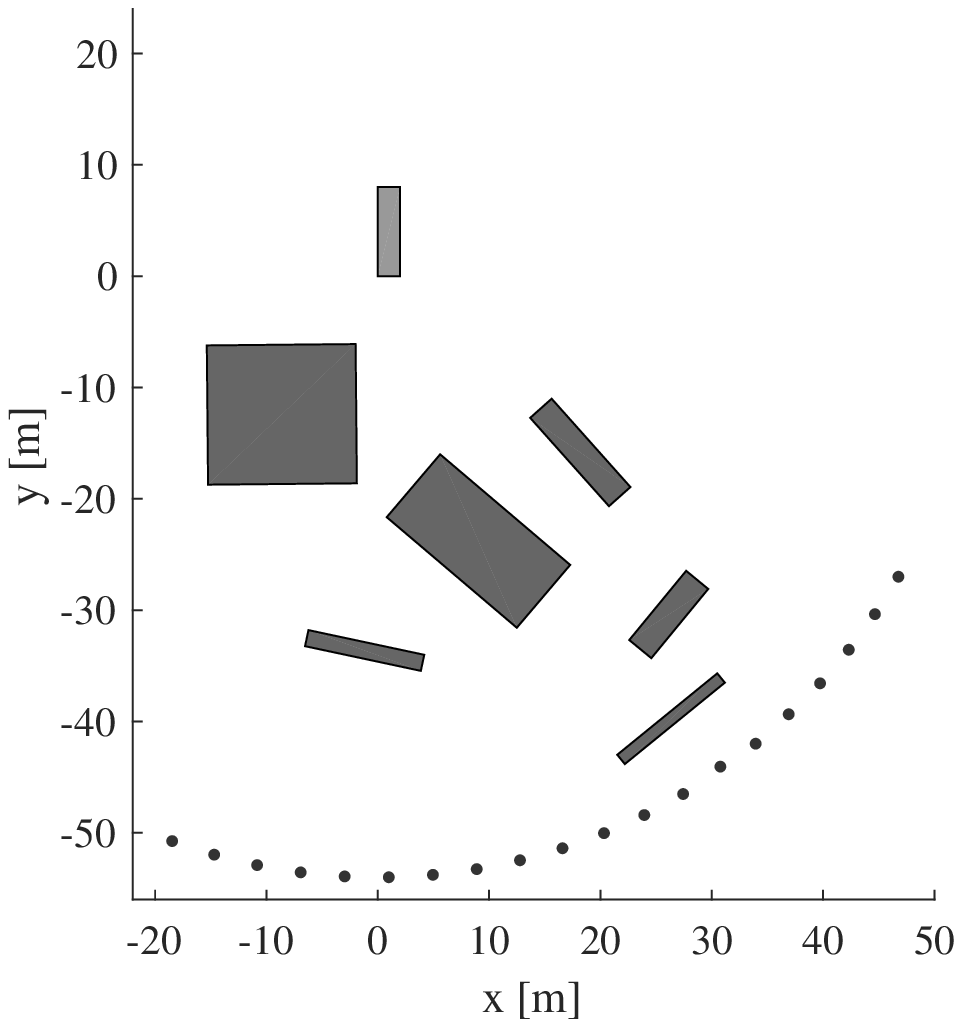}
\caption{Simulated task.}
\label{fig:simulationtask}
\end{subfigure}
\caption{Simulation system and task setup.}
\label{fig:simulationsetup}
\end{figure}

\subsection{Overview}

Fig. \ref{fig:simulationsetupblurlabelscolor} shows the experimental system that was setup to exercise and investigate human guidance behavior and the underlying control and perceptual processes. The first-person task is created in a simulated 3D environment \cite{feit2015experimental} that allows for precise control of available visual cues, a repeatable environment configuration, and consistent vehicle dynamic response. 

\subsubsection{Experimental Task}

During each task trial, the subject begins at one of 20 start positions and uses a controller to generate commands that move the vehicle to a common goal corridor. The objective in each task is to reach the goal in minimal time while avoiding obstacles. Travel time is displayed to the subject when they reach the goal. The subject is allowed to attempt the task multiple times from each start location until they feel that they have reached their best time. The 3D environment model consists of a flat ground plane and a narrow goal corridor that restricts vehicle direction when it reaches the goal.

\subsubsection{Data Collection}

A Tobii gaze tracking device is used to measure the gaze vector of the subjects during the task. Gaze data is processed to determine which portions of the environment the subject is focusing on and what perceptual functions are important during specific phases of the task. In addition to gaze, the system records vehicle position, velocity, and the subjects' control inputs.

\subsection{Agent-Environment System}
The system simulates an ideal unicycle-type vehicle response. The system is constrained to 2D motion, having two control inputs consisting of forward and lateral acceleration. The dynamics are defined in Eqn. \ref{eqn:vehicle_dynamics}.
\begin{eqnarray}
\begin{bmatrix}
\dot{x} \\ \dot{y} \\ \dot{\psi}
\end{bmatrix}
&=&
\begin{bmatrix}
v \cos \psi \\ v \sin \psi \\ \min(u_{lat}/v, \omega_{max})
\end{bmatrix} \label{eqn:vehicle_dynamics} \\
\dot{v} &=& k_{acc}*u_{lon} - k_{drag}*v \nonumber
\end{eqnarray}
In Eqn. \ref{eqn:vehicle_dynamics}, $u_{lat}$ is constrained based on lateral acceleration, $u_{lon} \in \left[0, u_{max}\right]$, and $k_{drag}$ is a drag coefficient that provides speed stability. $k_{acc}$ is an acceleration gain that defines the maximum vehicle speed. The resulting system is easily controlled, but incorporates essential constraints present in real vehicles. The lateral acceleration limit reproduces either a limit on aircraft bank angle, or a limit on lateral traction of a wheeled vehicle on slippery surfaces, and requires the subject to plan turns in advance to successfully navigate the course.

\begin{figure}[t]
\begin{subfigure}[t]{0.42\textwidth}
\centering
\includegraphics[width=0.95\linewidth]{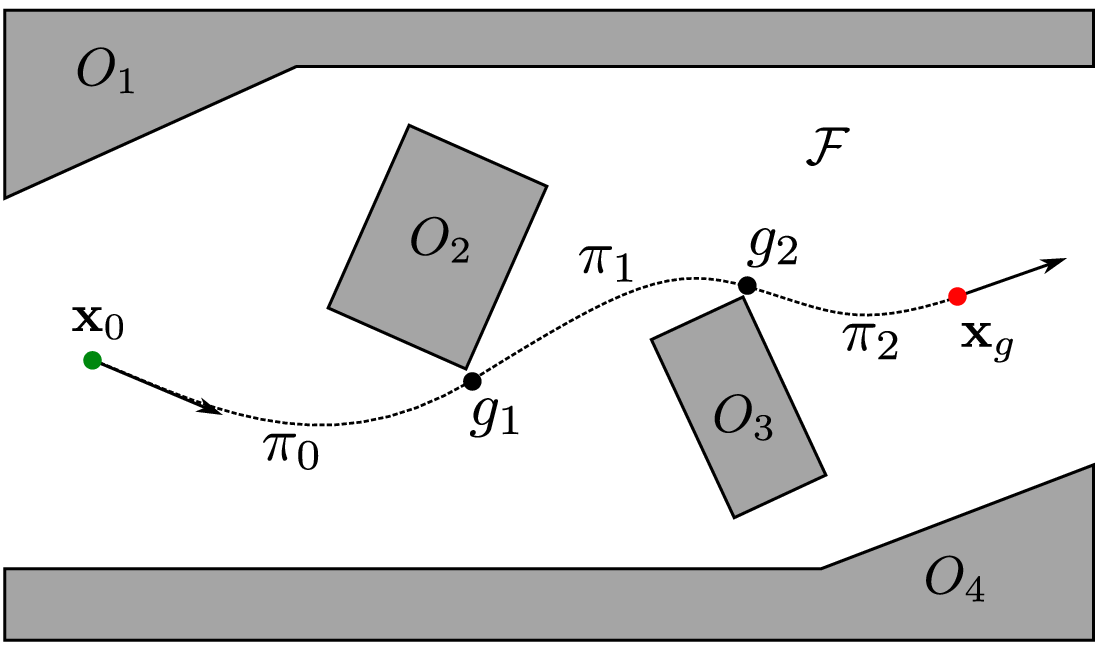}
\caption{Guidance problem.}
\label{fig:problem_overview}
\end{subfigure}
\hfil
\begin{subfigure}[t]{0.48\textwidth}
\centering
\includegraphics[width=0.95\linewidth]{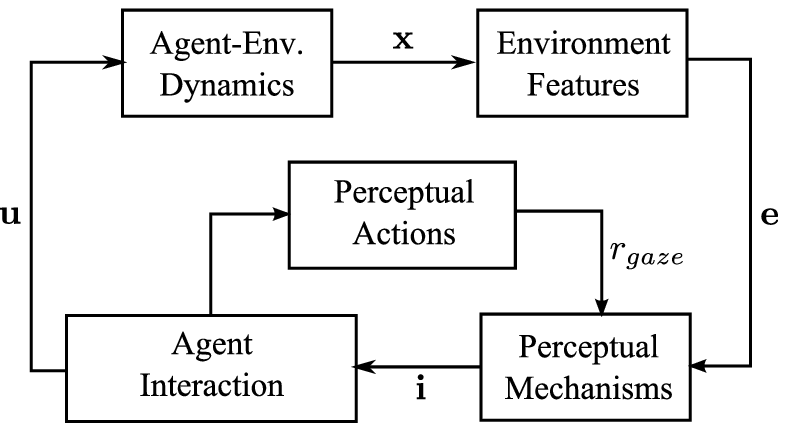}
\caption{Agent-environment system.}
\label{fig:guidanceframework2}
\end{subfigure}
\caption{Agent-environment system.}
\label{fig:agentenvironmentsystem}
\end{figure}

The human guidance task can be formally defined as a constrained optimal control problem as illustrated in Fig. \ref{fig:problem_overview}, consisting of a  system model over a state-space domain, $\mathcal{X} \subset \mathbf{R}^n$. Motion of the system is constrained by the dynamics of the vehicle or body, $\dot{\mathbf{x}}(t) = f(\mathbf{x}(t), \mathbf{u}(t)$, with control input sequence $\mathbf{u}(t) \in \mathcal{U} \subset \mathbf{R}^m$, for time $t \in \left[t_0,t_f\right] \subset \mathbf{R}$. The workspace, $\mathcal{W} \in \mathcal{X}$ is a system subspace defining the spatial domain of the vehicle, and includes free space ($\mathcal{F}$) and environment objects ($\mathcal{O}_E = \bigcup_{i=1}^{k} O_i$), such that $\mathcal{W} = \mathcal{O}_E \cup \mathcal{F}$. Environment objects $\mathcal{O}_E \in \mathcal{W}$ are discrete, polygonal obstacles constituting subsets of the workspace, $\{C_1, \dots, C_n \} \in \mathcal{O}_E$. The optimal control objective is to guide a system from an initial state, $\mathbf{x}_0 \in \mathcal{F}$ to a goal set, $\mathcal{X}_g \in \mathcal{F}$, while satisfying system dynamics and spatial constraints, and minimizing some trajectory cost function.

\section{Functional Behavior Model}

\begin{figure}[t]
\centering
\begin{tabular}{lp{6.5cm}}
\textbf{Symbol} & \textbf{Description} \\
\toprule
$\mathbf{x} \in \mathcal{X}$ & Agent state. \\
$\mathbf{e} \in \mathcal{E}$ & Environment state. \\
$\mathbf{i} \in \mathcal{I}$ & Environment information. \\
$\mathbf{u} \in \mathcal{U}$ & Control action. \\
$\mathbf{r}_{gaze} \in \mathcal{R}$ & Gaze direction. \\
\bottomrule
\end{tabular}
\caption{Agent-environment system signals.}
\label{tab:systemmodelsignals}
\end{figure}

The perception-action model presumes that agents are embedded in their environment, forming a closed-loop dynamic system, as shown in Fig. \ref{fig:guidanceframework2}. The environment state $\mathbf{e} \in \mathcal{E}$ describes environment objects $O_i$ relative to the agent. The function $\mathbf{e} = h(\mathbf{x})$ describes the relationship between agent state and environment state, consisting of a transformation into the agent's first-person reference frame. Perceptual mechanisms describe information $\mathbf{i}$ that an agent extracts from the environment state using a perceptual function, $\mathbf{i} = g(\mathbf{e}, \mathbf{r}_{gaze})$. Perceptual actions include $\mathbf{r}_{gaze}$, eye and head motion, which modulates the perceptual information the agent can obtain from the environment. 

In~\cite{mettler2015systems} closed-loop agent-environment dynamics are a hierarchical system. In this system, behavior is decomposed into subtasks but also into distinct control and perceptual modes. The agent interaction function describes how an agent chooses control actions $\mathbf{u}$ in response to perceptual information, $\mathbf{u} = k(\mathbf{i})$. Agent-environment dynamics describe how system state changes in response to control inputs, $\dot{\mathbf{x}} = f(\mathbf{x}, \mathbf{u})$. Both of these functions are partitioned into three hierarchical functional levels consisting of discrete task planning, guidance trajectory generation, and regulation of higher-order dynamics (following the hierarchical functional model described in \cite{mettler2013hierarchical}). The agent-environment interaction functions at each hierarchical level will be described in detail in the next section. The agent-environment system model is:
\begin{eqnarray}
\text{Task transition:}&  g_{k+1} = \Phi(g_k, \pi_k) \\
\text{Kinematics:}&  \dot{\mathbf{x}}_p(t) = \mathbf{v_{ref}}(t) \\
\text{Dynamics:}&  \dot{\mathbf{v}} = f(\mathbf{v},\mathbf{u})
\label{eqn:agentinteractionfunctions}
\end{eqnarray}
In Eqn. \ref{eqn:agentinteractionfunctions}, a subgoal $g_k \in \mathcal{X}$ is a task state representing an intermediate goal on the solution trajectory $\overleftarrow{s}_{x_g}$, as illustrated by points $\lbrace g_1, g_2 \rbrace$ in Fig. \ref{fig:problem_overview}. Transitions between subgoals are specified by $\Phi(g_k, \pi_k)$, in which $g_k$ is the current subgoal and $\pi_k$ is a guidance trajectory moving the system to subgoal $g_{k+1}$. A solution trajectory consists of a sequence of segments, $\overleftarrow{s} = \{\pi_1, \dots, \pi_n\}$. A guidance element $\pi_k$ is a path between start and end task states $\mathbf{x_p}(0)$ and $\mathbf{x}(T)$ as specified by a reference velocity, $\mathbf{v}_{ref}(t)$ for $t \in \left[0,T\right]$. Guidance elements $\pi_0$, $\pi_1$, and $\pi_2$ are labeled in Fig. \ref{fig:problem_overview}. Finally, vehicle velocity $\mathbf{v}(t)$ is modeled by dynamics $f(\mathbf{v}, \mathbf{u})$, evolving based in response to control inputs.

\subsection{First-Person Task Definition}

\begin{figure}[tbph]
\centering
\includegraphics[width=0.50\linewidth]{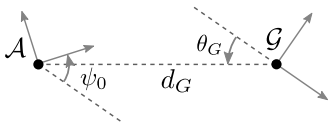}
\caption{First-person guidance geometry}
\label{fig:firstpersonguidancegeometry}
\end{figure}

Fig. \ref{fig:firstpersonguidancegeometry} illustrates two reference frames used to formulate the human guidance task. Reference frame $\mathcal{A}$ is centered at the agent position, and is the first-person frame from which visual cues are perceived. $\mathcal{G}$ is centered at the goal location, and oriented with the velocity vector at the goal state. The agent must understand how $\mathcal{A}$ relates to $\mathcal{G}$ to relate perceived cues in the visual field to relevant goal information, and to generate spatial guidance behavior. In the experimental task presented here, the workspace is a 2D, $\mathcal{W} \in \mathbf{R}^2$. The agent position $\mathbf{x}_p \in \mathcal{W}$ is expressed in polar coordinates as $\mathbf{x}_G = \left[\theta_G \,\, d_G \right]$ and the agent initial orientation expressed as $\psi_0$, as illustrated in Fig. \ref{fig:firstpersonguidancegeometry}.

\section{Guidance Behavior Decomposition}

\subsection{Overview}

The behavior decomposition is based on two hypotheses about human dynamic spatial behavior: first, that they use a finite number of dynamic control modes to achieve the range of guidance behavior necessary to perform a task.  Further, that guidance behaviors display functional equivalence, i.e., mode transitions occur following a specific sequence of sensory and control actions~\cite{mettler2017emergent}. Such behaviors can be modeled using a finite-state automaton \cite{mettler2010agile, frazzoli2005maneuver}. The second hypothesis is that dynamic control modes occur within lower-dimension subspaces of the agent-environment state-space. The dimension of the control manifold can be reduced using different approaches. For example, defining meta-controls \cite{braun2010structure}, using latent states corresponding to a hierarchical task representation \cite{mettler2013hierarchical}, or with perceptual guidance principles such as Tau theory~\cite{lee1976theory}. 

Behavior data decomposition consists of two steps: first dividing behavioral data into functionally equivalent segments and second, clustering the segments into a small number of dynamic control modes that describe the majority of subject behavior. Behavior segments are functionally equivalent if they belong to the same subspace manifold. In this work, functional equivalence between segments is characterized by the set of active constraints and the statistical relationships between unconstrained signals.

\begin{figure*}[t]
\centering
\begin{subfigure}{0.38\textwidth}
\centering
\includegraphics[height=2.9cm]{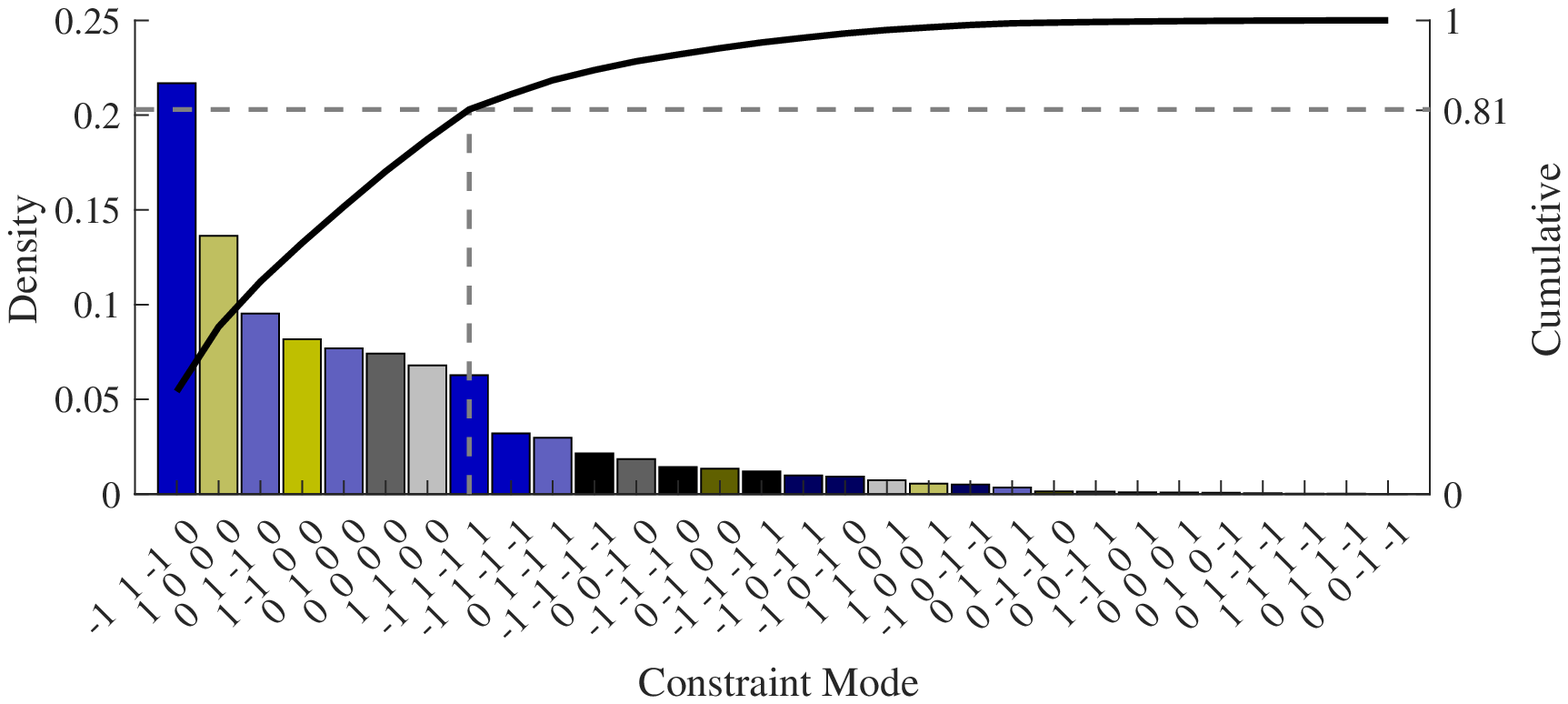}
\caption{Relative density of constraint classes.}
\label{fig:constraintclasshist}
\end{subfigure}
\hfil
\begin{subfigure}{0.36\textwidth}
\centering
\includegraphics[height=2.9cm]{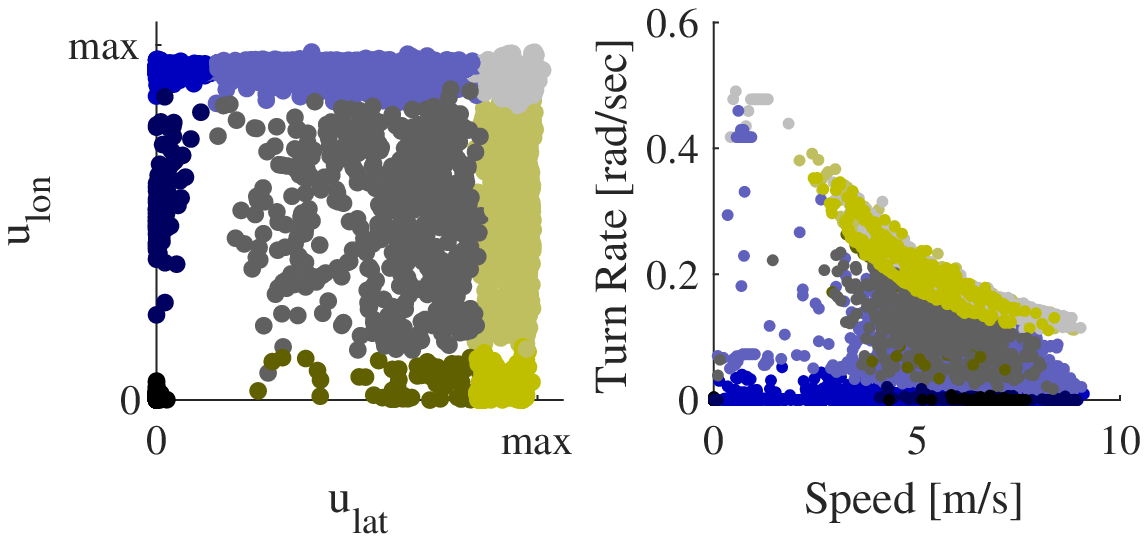}
\caption{Control and dynamic state domains.}
\label{fig:constraintclassscatter}
\end{subfigure}
\hfil
\begin{subfigure}{0.22\textwidth}
\centering
\includegraphics[height=2.9cm]{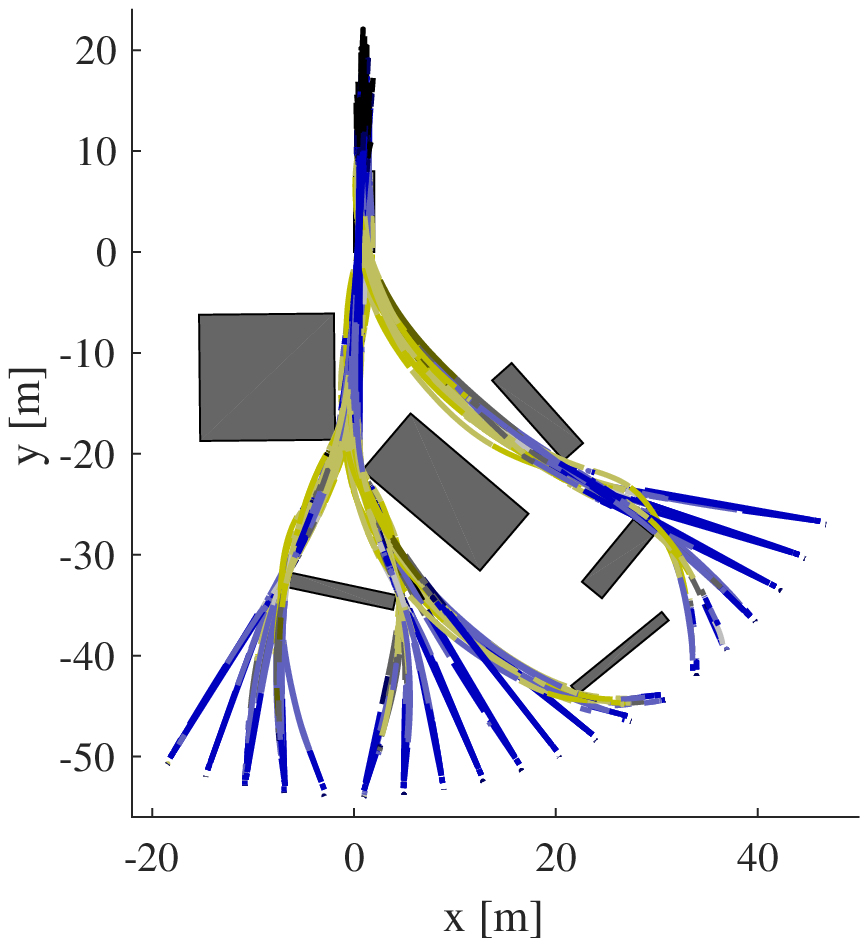}
\caption{Spatial domain.}
\label{fig:constraintclasstrajectories}
\end{subfigure}
\caption{Constraint class decomposition.}
\label{fig:constraintclasses}
\end{figure*}

\subsection{Constraint Class Identification}

The dynamic state of the system is defined as the product set of control inputs, $\lbrace u_{lat}, u_{lon} \rbrace \in \mathcal{U}$ and rates, $\lbrace v, \omega \rbrace \in \mathcal{V}$, $\mathcal{X}_d = \mathcal{V} \times \mathcal{U} = \lbrace u_{lat}, u_{lon}, v, \omega \rbrace$. The dynamic constraint state for a trajectory point is $\mathbf{c}(\mathbf{x}_d) = \left[ c(u_{lat})), c(u_{lon}), c(\omega) c(v) \right]$, where $c(x)$ is the constraint state of an individual parameter value $x$ and is defined as:
\begin{equation}
c_i(x)= 
\begin{cases}
    1,& \text{if } x - x^{max} \approx 0\\
    -1,& \text{if } x^{min} - x \approx 0\\
    0,& \text{otherwise}
\end{cases}
\label{eqn:constraintstate}
\end{equation}
Three values are used ($\lbrace -1, 0, 1 \rbrace$) to explicitly account for minimum and maximum constraint activation for a signal being mutually exclusive.

Fig. \ref{fig:constraintclassscatter} shows a scatter of trajectory points in both a control input and dynamic state domains. Color in both plots indicates the constraint state of the control inputs for that point ($u_{lat}$ in yellow and $u_{lon}$ in blue). Fig. \ref{fig:constraintclasshist} shows the relative frequency for each of the 30 constraint mode clusters. This constraint state definition transforms a trajectory $\mathbf{x}(t)$ for $t \in \left[0,T\right]$ to a constraint-state trajectory, $\mathbf{c}(t) \in \lbrace -1, 0, 1 \rbrace^4 \times T$. Fig. \ref{fig:constraintclasstrajectories} shows the constraint state along each recorded trajectory.

\subsection{Spatial Subgoal Identification}

\begin{figure}
\centering
\begin{subfigure}[t]{0.4\textwidth}
\centering
\includegraphics[height=3.7cm]{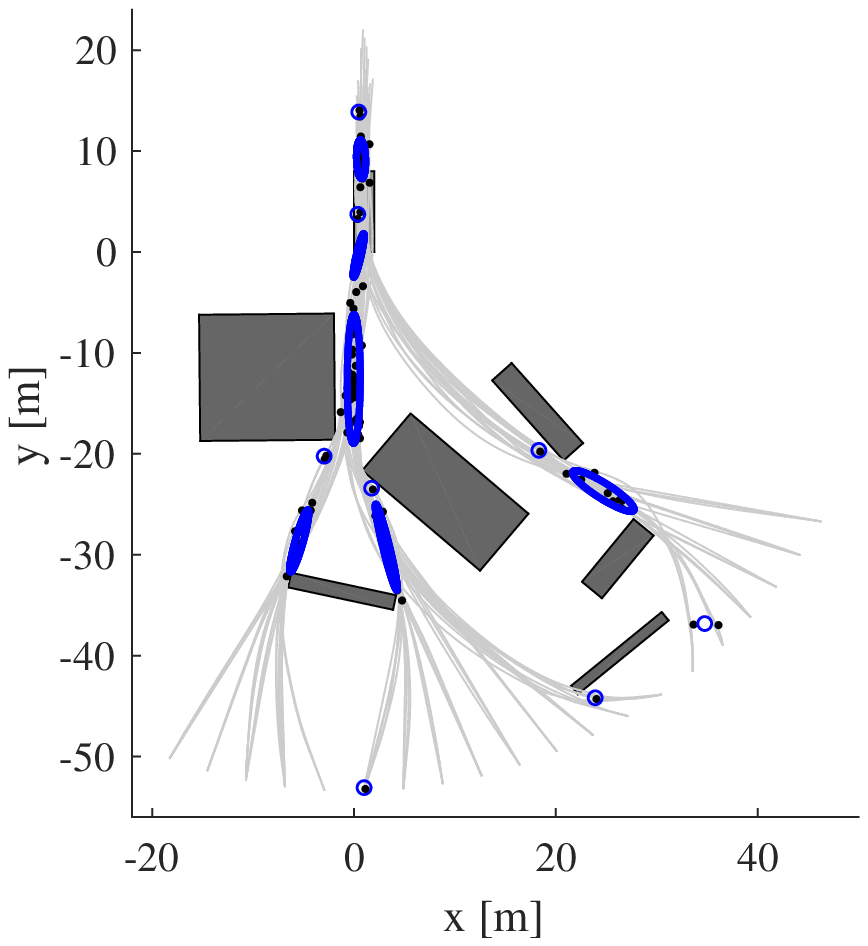}
\caption{Subgoal clusters.}
\label{fig:dynamicsubgoalscatter}
\end{subfigure}
\hfil
\begin{subfigure}[t]{0.27\textwidth}
\centering
\includegraphics[height=3.7cm]{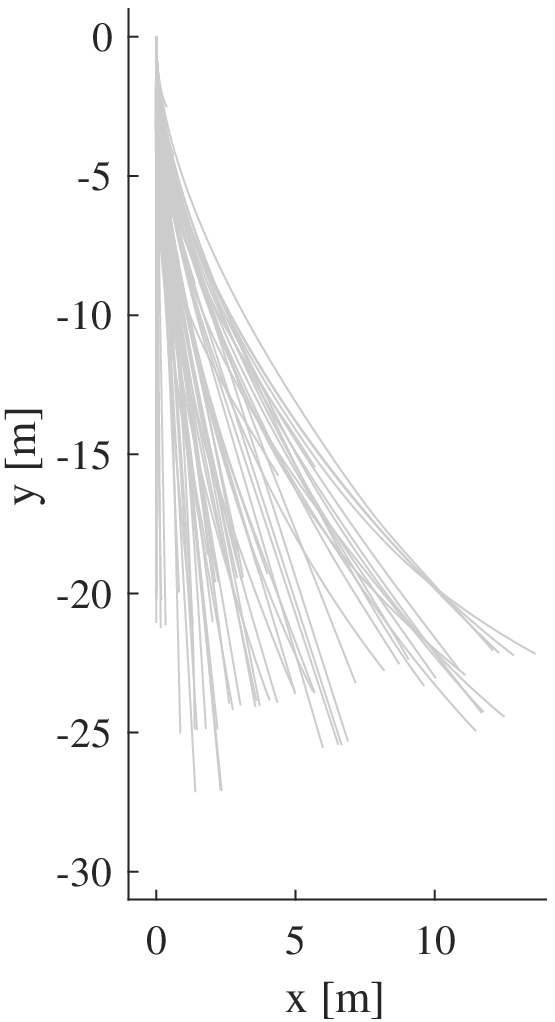}
\caption{Segments.}
\label{fig:dynamicsegements}
\end{subfigure}
\hfil
\begin{subfigure}[t]{0.27\textwidth}
\centering
\includegraphics[height=3.7cm]{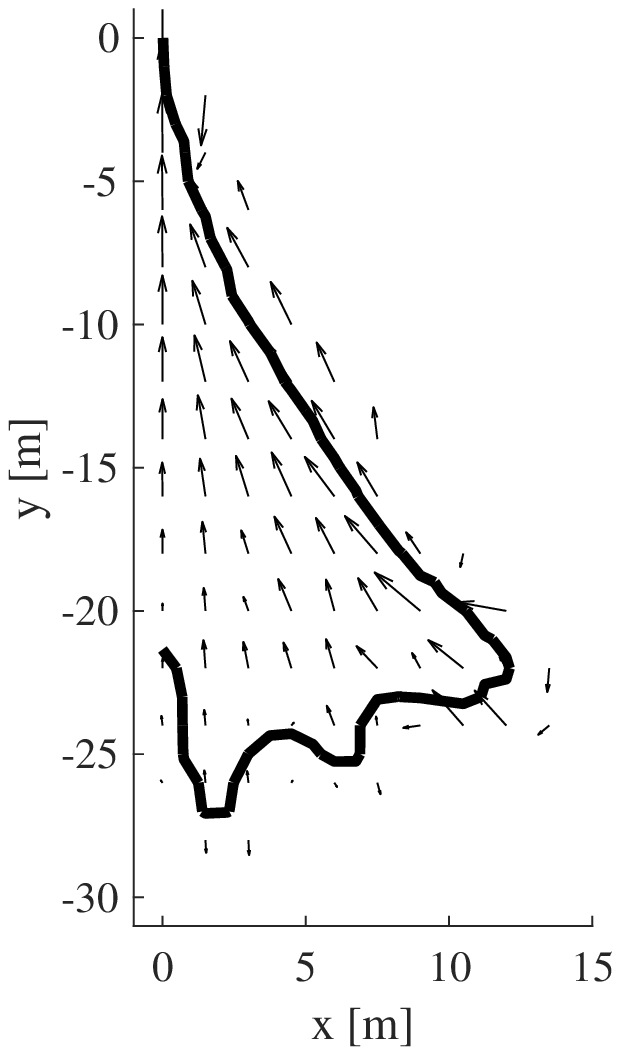}
\caption{VVF.}
\label{fig:gpsvfvvf}
\end{subfigure}
\caption{Planning-level behavior decomposition.}
\label{fig:planningdecomposition}
\end{figure}

Subgoals are important for task planning because they divide the global problem into a set of equivalent local guidance problems. Subgoals in prior work were formalized based on the concept of causal state, which defines the subgoal equivalence used for their identification~\cite{kong2013modeling}. This type of subgoal can also be described using the principle of optimality \cite{feit2015human, feit2017subgoal}. From a functional standpoint, they represent transitions between regions that share similar active spatial constraints. This property is less restrictive and relevant for global behavior that hasn't yet converged to some optimal or near-optimal behavior~\cite{mettler2017emergent}.

A different approach to subgoal identification is based on the concept of relevant goal information \cite{dijk2011grounding} and MDP policy options \cite{sutton1999mdps}. During guidance, motion can be described as a sequence of perceptual gap closures in the sense of Tau-theory \cite{lee1976theory, lee1998guiding}. Reaching a subgoal corresponds to gap closures, and results in a discontinuity in control actions.

In the present study, subgoal locations are estimated using functional properties from constraint class transitions along each trajectory. The hypothesis is that subgoals occur at the transitions from turning to rectilinear motion, because at this transition, $c(u_{lat}) = -1$ becomes satisfied. The subgoal equivalence predicts that these constraint transition points correspond to a set of subgoals that are invariant with respect to trajectory start location. Fig. \ref{fig:dynamicsubgoalscatter} shows that subgoal candidate points form concentric clusters, and occur in areas where multiple trajectories come together.

\subsection{Guidance Behavior and Control Mode Identification}

The identified subgoals divide trajectories into segments of guidance behavior. Each segment is an example of continuous, goal-reaching motion in response to perceptual information. The extracted segments are aggregated through rigid-body translation and rotation Fig. \ref{fig:dynamicsegements} shows the resulting segments, with goal velocity in the positive $y$-axis. 

In previous work, this aggregate set of guidance behavior was used to estimate a spatial cost-to-go function \cite{mettler2013mapping} and also approximated using Gaussian Process regression \cite{feit2016extraction} as shown in Fig. \ref{fig:gpsvfvvf}. In the present work, the aggregate set is decomposed into a repertoire of control modes $\mathcal{M}$ based on functional characteristics. 

The individual segments are then used to identify the underlying perceptual and control modes and associated mechanisms. Control modes are identified by clustering constraint modes that are functionally similar, in terms active constraints and relationships between free variables. The identified modes are then labeled, for example ``turning" or ``rectilinear", to produce semantic mode describing the meaning of these actions in the context of the guidance task.

\subsubsection{Graphical Modeling}

\begin{figure}[tbph]
\centering
\includegraphics[width=0.95\linewidth]{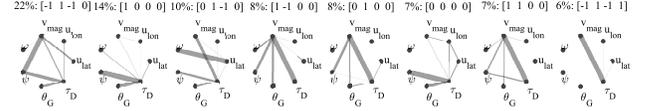}
\caption{Sparse MRFs depicting signal relationships within first eight constraint classes and their use frequency in \%.}
\label{fig:constraintclassgraphs}
\end{figure}

Graphical modeling is used to determine signal relationships present within each constraint mode. These relationships include known dynamic and kinematic relationships in the vehicle-environment model, as well as perceptual guidance relationships implemented by the human operator. The joint probability distribution between guidance signals is modeled as a Markov random field (MRF), with graph edge weights describing the inverse covariance of the adjacent random variables. A sparse MRF is estimated using the Matlab SLEP package \cite{liu2009slep}. Fig. \ref{fig:constraintclassgraphs} depicts graphical models describing the eight most common constraint modes.

\begin{figure*}
\centering
\begin{subfigure}{0.39\textwidth}
\centering
\includegraphics[height=2.5cm]{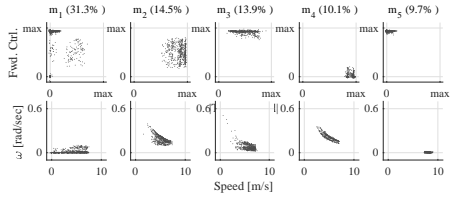}
\caption{Semantic mode point scatters.}
\label{fig:modeclusterscatters}
\end{subfigure}
\hfil
\begin{subfigure}{0.39\textwidth}
\centering
\includegraphics[height=1.8cm]{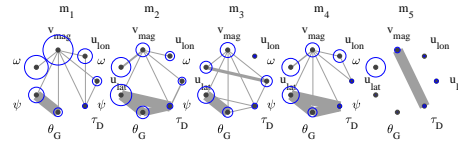}
\vspace{7mm}
\caption{Semantic mode graphical models.}
\label{fig:modeclustergraphs}
\end{subfigure}
\hfil
\begin{subfigure}{0.19\textwidth}
\centering
\includegraphics[height=2.0cm]{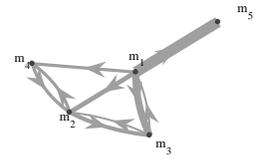}
\vspace{5mm}
\caption{Mode transition graph.}
\label{fig:modetransitiongraph}
\end{subfigure}
\vfil
\vspace{2mm}
\begin{subfigure}[t]{0.9\textwidth}
\centering
\includegraphics[height=2.2cm]{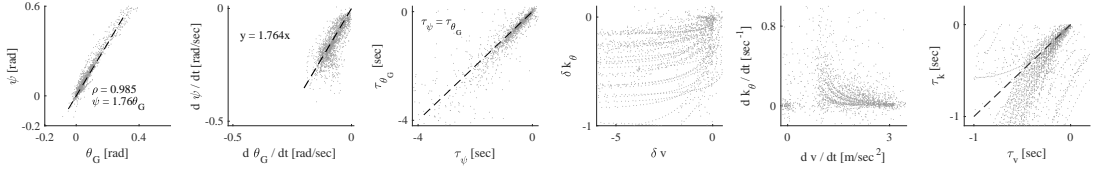}
\caption{Turning (left three plots) and Rectilinear (right three plots) behavior.}
\label{fig:tausteering2}
\end{subfigure}
\caption{Semantic mode clusters and Tau coupling.}
\label{fig:modeclusters}
\end{figure*}

\subsubsection{Control Mode Clusters}

Next, constraint classes are clustered to identify a small number of semantic control modes. Class similarity is formulated as a combination of graph edge similarity and constraint mode similarity. Edge similarity is quantified as the Jaccard index of graph edge sets. Constraint mode similarity is the $L_1$-norm of the difference between constraint states:
\begin{equation}
S_m(i,j) = min(E_i, E_j) / max(E_i, E_j) - w*||c_i - c_j||_1
\label{eqn:similaritymetric}
\end{equation} 
Eqn. \ref{eqn:similaritymetric} is the similarity metric, with weight $w$ defining the trade-off between edge similarity and constraint mode similarity. Constraint modes are clustered using this metric, resulting in five control modes as shown in Fig. (\ref{fig:modeclusters}).

\subsubsection{Control Mode Transitions}

The hypothesis is that motion behavior is based on an ideal transition mode sequence that is common across guidance segments, and for which observed mode sequences are a noisy measurement. Observed mode transitions probabilities are depicted as a graph in Fig. \ref{fig:modetransitiongraph}. The actual mode transition sequence is considered as a hidden Markov model (HMM) from which the most-likely sequence of modes is estimated using the Viterbi algorithm, with a transition probability matrix $T \in \mathbf{R}^{5x5}$ and a measurement model, $Z \in \mathbf{R}^{5x30}$ specifying the likelihood of observing each of 30 constraint classes within each control mode.

\subsection{Control Modalities}

\begin{table}
\centering
\begin{tabular}{p{2.3cm}|p{2.5cm}|p{2.3cm}}
\centering
Modality 
& 
\centering
Relationships 
& 
{
\centering
Description
}
\\
\toprule
\hline
\vspace{0mm}
\begin{minipage}{.13\textwidth}
\centering
\includegraphics[height=1.2cm]{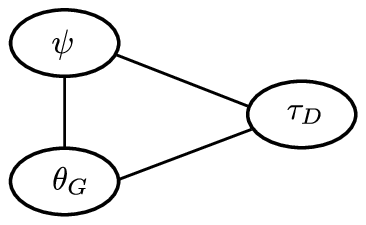}
\end{minipage}
\vspace{0.5mm}
&
\centering
\vspace{0mm}
$\begin{aligned}
\tau_{\psi} &= k_{\psi} \tau_d \\
\tau_{\theta} &= k_{\theta} \tau_d \\
\psi_{ref} &= k_{\theta} \theta_G
\end{aligned}$ & 
\vspace{0.5mm}
\begin{tabular}{rl}
Type:&Agent \\
\multicolumn{2}{c}{$\lbrace m_2, m_3, m_4 \rbrace$}
\end{tabular}
\\
\hline
\vspace{1mm}
\begin{minipage}{.13\textwidth}
\centering
\includegraphics[height=1.08cm]{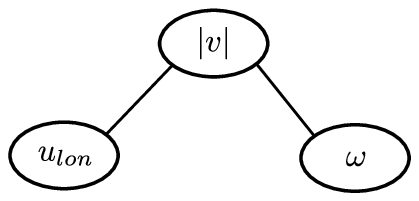}
\end{minipage}
\vspace{1mm}
& 
\centering
\vspace{1mm}
$\begin{aligned}
u_{lon} &= v_{ref} \\
+ k(v &- v_{ref}) \\
v_{ref} =& a_y^{max} / \omega_{ref}
\end{aligned}$
\vspace{1mm}
& 
\vspace{1mm}
\begin{tabular}{r l}
Type: & System \\
\multicolumn{2}{c}{$\lbrace m_2, m_4 \rbrace$}
\end{tabular}
\\
\hline
\vspace{1mm}
\begin{minipage}{.13\textwidth}
\centering
\includegraphics[height=0.42cm]{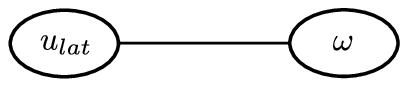}
\end{minipage}
\vspace{1mm}
& 
\centering
\vspace{1mm}
$u_{lat} = \omega_{ref}$
& 
\vspace{1mm}
\begin{tabular}{r l}
Type: & System \\
\multicolumn{2}{c}{$\lbrace m_3, m_4 \rbrace$}
\end{tabular}
\\
\hline
\vspace{1mm}
\begin{minipage}{.13\textwidth}
\centering
\includegraphics[height=0.42cm]{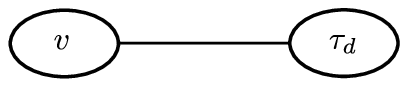}
\end{minipage}
\vspace{1mm}
&
\centering
\vspace{1mm}
$\begin{aligned}
\tau_v &= k_v \tau_d \\
\tau_k &= k_k \tau_d \\
\delta v &= k_b \delta k_{\theta}
\end{aligned}
$
\vspace{1mm} 
&
\vspace{1mm}
\begin{tabular}{r l}
Type: & Agent \\
\multicolumn{2}{c}{$\lbrace m_1, m_5 \rbrace $}
\end{tabular}
\\
\bottomrule
\end{tabular}
\caption{Perceptual guidance modalities.}
\label{tab:controlmodalities}
\end{table}

Next, groups of related signals that indicate perceptual guidance relationships are identified in each control mode. Control modes contain common related signal groups, or control modalities, which are of two types: vehicle dynamic modalities that are defined by the vehicle system model, and agent control modalities. Agent control modalities represent perceptual guidance relationships used by the operator to generate guidance behavior.

Table \ref{tab:controlmodalities} describes four control modalities: steering, speed-turnrate, a turnrate-control, and braking-control. The first and last modalities represent perceptual guidance relationships that are implemented by the operator. The second and third are system dynamic relationships, indicating which constraints are currently active. 

\subsection{Tau Model of Mode Transitions}

\begin{figure}[t]
\centering
\includegraphics[height=2.5cm]{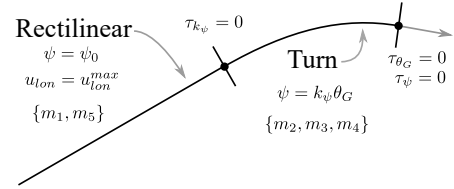}
\caption{Control mode sequence.}
\label{fig:controlmodetransitiondiagram}
\end{figure}

The observed control mode transition sequence suggests that guidance behavior can be described as a series of gap closures consistent with Tau theory \cite{lee1976theory}. Control modes typically proceed in the order: Rectilinear $\rightarrow$ Turn as depicted in Fig. \ref{fig:controlmodetransitiondiagram}. 

\subsubsection{Turning Model}
During turns, vehicle heading converges to the subgoal heading ($\psi \rightarrow \psi_{G}$), the goal bearing error converges to zero ($\theta_G \rightarrow 0$) and the distance to the subgoal converges to zero ($d \rightarrow 0$).

The multiple gap closures such as this can be coordinated by coupling the Tau of the gaps, i.e. $\tau_A = k \tau_B$ \cite{lee1998guiding}. This coupling predicts that turning behavior is guided by $\tau_{\psi} = k \tau_{\theta_G}$, where $\tau_{\psi} = \psi / \dot{\psi}$ and $\tau_{\theta_G} = \theta_G / \dot{\theta_G}$. The left three plots in Fig. \ref{fig:tausteering2} illustrate turning Tau relationships. The third plot shows that $\tau_{\psi}$ and $\tau{\theta_G}$ converge to a linear relationship as they approach zero, consistent with Tau-coupling. This coupling results in linear relationships $\psi \approx 1.76 \theta_G$, as shown in the leftmost plot.

\subsubsection{Rectilinear Model}
Prior to starting a turn, vehicle heading is constant, while goal bearing error is decreasing. To determine when turning begins, the agent may use the steering ratio gap. During rectilinear motion, $k_{\theta}$ is decreasing until it reaches the desired constant value for the turning mode. This change in steering ratio prior to initiating the turn can be modeled as a gap closure indicating to the agent when to initiate the turn.
\begin{eqnarray}
\delta k =& (k - k_{steer}) \nonumber \\
\tau_k =& \delta k / \delta \dot{k}
\end{eqnarray}

\section{Perceptual Behavior Decomposition}

The gaze-based approach to perceptual behavior decomposition identifies gaze patterns based on gaze measurements. Gaze behavior is classified into the typical motion types: fixations, smooth-pursuits, and saccades \cite{hayhoe2005eye}. Prior work studying third-person motion guidance shows that eye motion is associated with two perceptual functional modes: measuring a gap to the goal, and tracking current vehicle position \cite{andersh2014modeling}. 

\subsection{Gaze Classification}

\begin{figure*}[t]
\begin{subfigure}[t]{0.40\textwidth}
\centering
\includegraphics[height=5.0cm]{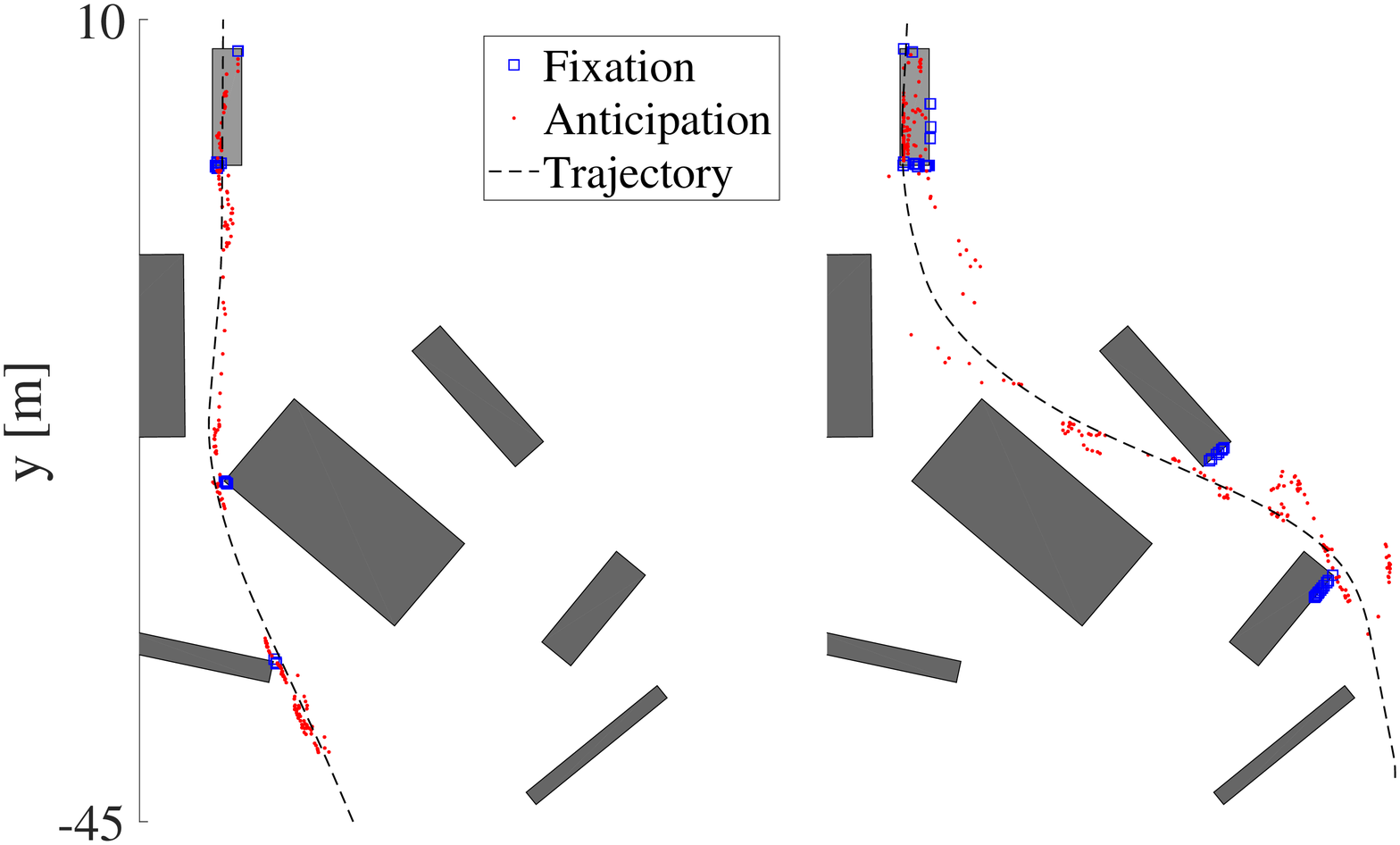}
\caption{Smooth-pursuit gaze behavior clusters.}
\label{fig:gazeclasssegmentexamples}
\end{subfigure}
\centering
\begin{subfigure}[t]{0.17\textwidth}
\centering
\includegraphics[height=4.5cm]{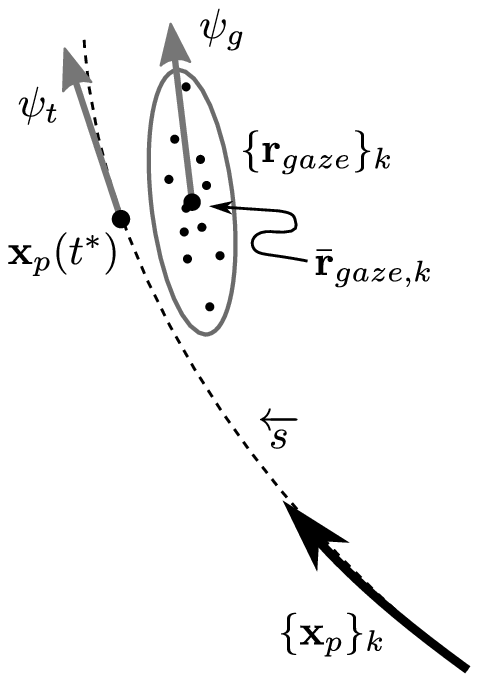}
\caption{Gaze heading.}
\label{fig:gazeanticheadingdiagram}
\end{subfigure}
\hfil
\begin{subfigure}[t]{0.22\textwidth}
\centering
\includegraphics[height=5.0cm]{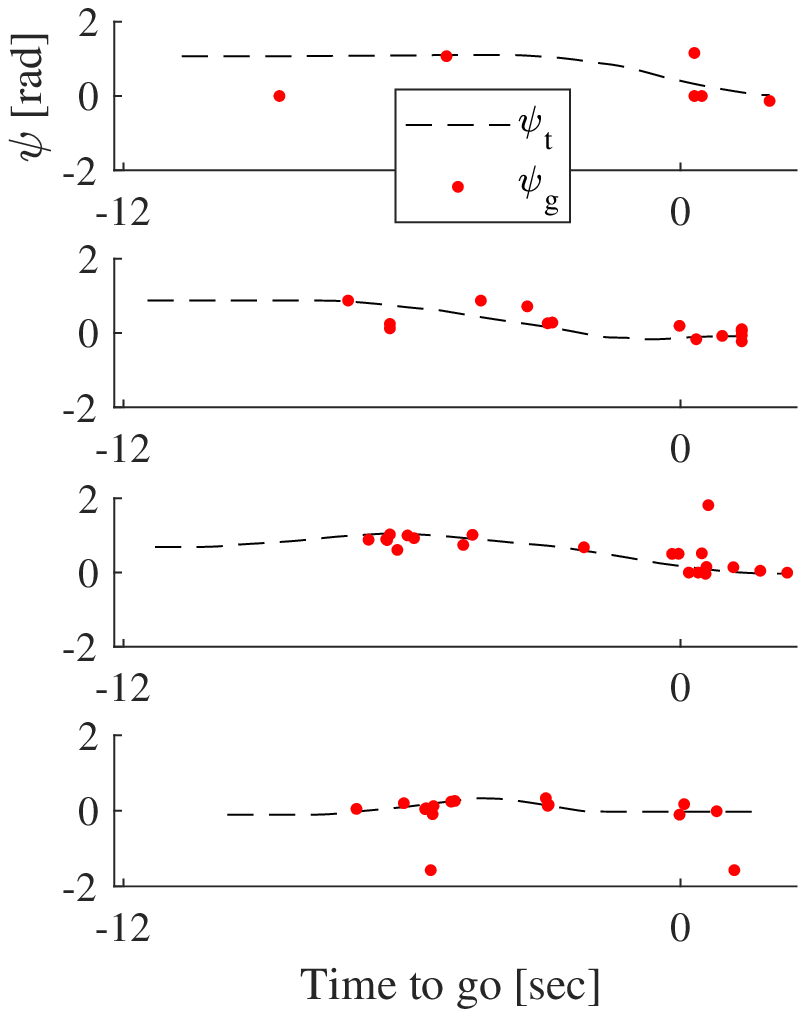}
\caption{$\psi_g$ and $\psi_t$ vs. time.}
\label{fig:gazeanticheading}
\end{subfigure}
\hfil
\begin{subfigure}[t]{0.17\textwidth}
\centering
\includegraphics[height=5.0cm]{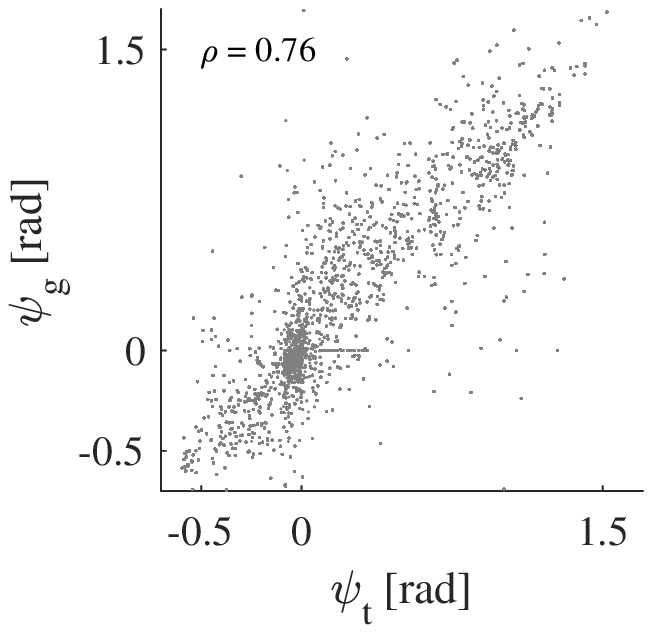}
\caption{$\psi_g$ vs. $\psi_t$.}
\label{fig:headingcorrelation}
\end{subfigure}
\caption{Anticipatory Gaze Behavior.}
\label{fig:anticipatorygazefunction}
\end{figure*}

Subject gaze behavior is first decomposed into the three modes using a Markov-based classification algorithm \cite{li2015classification}, which produces a sequence of gaze motion segments, separated by saccades: $\mathcal{P} = \lbrace p_1, \dots, p_n \rbrace$ as shown in Fig. \ref{fig:gazeclasssegmentexamples}.

\subsection{First-person Gaze Functions}

From Fig. \ref{fig:gazeclasssegmentexamples} it is apparent that segments of gaze behavior attend to different features, suggesting that they serve different functions to the agent. The hypothesis for first-person experiments is that gaze segments serve two primary perceptual functions: cue-fixation and trajectory anticipation. Gaze points are classified as cue-fixation or anticipation based on the distance from an obstacle. Cue-fixation segments coincide with environment obstacle boundaries. Trajectory anticipation gaze segments do not attend to obstacle boundaries, and appear to project along the agent's future path. Fig. \ref{fig:gazeclasssegmentexamples} illustrates two representative trajectories showing the gaze points mapped onto the 2D workspace.

To evaluate how well trajectory anticipation gaze predicts future agent behavior, the principal covariance direction of each gaze cluster is compared with the velocity direction at the nearest corresponding point along each trajectory. Fig. \ref{fig:gazeanticheading} shows the heading time-history of three example trajectories, along with the heading of each gaze cluster. Fig. \ref{fig:headingcorrelation} shows the correlation ($\rho = 0.76$) across all trials for this subject, suggesting that subjects use a predictive function to anticipate the future trajectory. A key point is that anticipatory gaze motion requires the agent to transform points in their visual field into the global task domain. This gaze behavior also demonstrates that trained operators have an accurate predictive model of guidance behavior.

\section{Conclusions}

\subsection{Specific Insights}

This analysis validates specific hypotheses on human guidance behavior: first, that humans learn and use a small set of guidance modes to support necessary environment interaction. Results show that the majority of observed behaviors can classified as one of five control modes. These results extend previous work that classified behavior using piecewise affine identification \cite{li2015towards} by showing that control modes can be defined by the set of active constraints. The second insight is that gaze motion during a guidance task can be classified into trajectory anticipation and cue fixation perceptual functions. Each function provides perceptual information to the agent, and defines a gaze motion profile. This extends work that previously identified similar perceptual functions in third-person helicopter control \cite{andersh2014modeling}. 

\subsection{Applications}

The understanding provided by the decomposition and modeling of human interaction patterns are relevant to applications in human-machine systems and autonomous motion guidance systems. First, this decomposition suggests an autonomous guidance approach that consists of deploying sequences of control modes. Each control mode may be selected based on specific task requirements. Second, perceptual guidance strategies, by providing a link between perceived quantities and actions, provide a means to evaluate and trade-off risk vs. performance. Finally, perceptual guidance strategies can be used as a perceptual filter to identify features in the environment most important to the current task and filter out irrelevant background information. These approaches would allow an autonomous system to transparently share information with a human operator, increasing situational awareness and allowing more natural interaction than existing approaches.

\subsection{Future Work}

This work identifies perceptual guidance relationships and mode switching sequences in human behavior, and suggests that these perceptual guidance modes may optimize either utility or an informational quantity such as empowerment, linking them to perception. Future work will investigate if the identified perceptual guidance relationships do satisfy or optimize some expected human motivational factors. 

The present work identified two perceptual gaze functions, but future work is required to understand the switching process between these functions. The experiments in the present work require subject to acquire only basic visual information to perform successfully. A topic of future work will be to understanding how attention is focused to relevant guidance information during higher-performance tasks in natural cluttered task environments.

\section*{Acknowledgments}
This work is financially supported by the National Science Foundation (CAREER 2013-18 CMMI-1254906).

\bibliographystyle{ieeetr}
\bibliography{behavior_decomp_references_arxiv}

\begin{thebibliography}{10}

\bibitem{bojarski2016end}
M.~Bojarski, D.~Del~Testa, D.~Dworakowski, B.~Firner, B.~Flepp, P.~Goyal, L.~D.
  Jackel, M.~Monfort, U.~Muller, J.~Zhang, {\em et~al.}, ``End to end learning
  for self-driving cars,'' {\em arXiv preprint arXiv:1604.07316}, 2016.

\bibitem{mettler2015systems}
B.~Mettler, Z.~Kong, B.~Li, and J.~Andersh, ``Systems view on spatial planning
  and perception based on invariants in agent-environment dynamics,'' {\em
  Frontiers in Neuroscience}, vol.~8, p.~439, 2015.

\bibitem{mettler2013hierarchical}
B.~Mettler and Z.~Kong, ``Hierarchical model of human guidance performance
  based on interaction patterns in behavior,'' {\em arXiv preprint
  arXiv:1311.3672}, 2013.

\bibitem{kong2013modeling}
Z.~Kong and B.~Mettler, ``Modeling human guidance behavior based on patterns in
  agent--environment interactions,'' {\em Human-Machine Systems, IEEE
  Transactions on}, vol.~43, no.~4, pp.~371--384, 2013.

\bibitem{mettler2013mapping}
B.~Mettler and Z.~Kong, ``Mapping and analysis of human guidance performance
  from trajectory ensembles,'' {\em Human-Machine Systems, IEEE Transactions
  on}, vol.~43, no.~1, pp.~32--45, 2013.

\bibitem{feit2016extraction}
A.~Feit and B.~Mettler, ``Extraction and deployment of human guidance
  policies,'' in {\em Submitted}, 2016.

\bibitem{feit2017subgoal}
A.~Feit and B.~Mettler, ``Subgoal planning algorithm for autonomous vehicle
  guidance,'' 2017.

\bibitem{feit2015human}
A.~Feit, A.~Verma, and B.~Mettler, ``A human-inspired subgoal-based approach to
  constrained optimal control,'' in {\em Decision and Control (CDC), 2015 IEEE
  54th Annual Conference on}, pp.~1289--1296, IEEE, 2015.

\bibitem{lee1976theory}
D.~N. Lee {\em et~al.}, ``A theory of visual control of braking based on
  information about time-to-collision,'' {\em Perception}, vol.~5, no.~4,
  pp.~437--459, 1976.

\bibitem{warren2006dynamics}
W.~H. Warren, ``The dynamics of perception and action.,'' {\em Psychological
  review}, vol.~113, no.~2, p.~358, 2006.

\bibitem{tishby2011information}
N.~Tishby and D.~Polani, ``Information theory of decisions and actions,'' in
  {\em Perception-action cycle}, pp.~601--636, Springer, 2011.

\bibitem{gavrilets2004human}
V.~Gavrilets, B.~Mettler, and E.~Feron, ``Human-inspired control logic for
  automated maneuvering of miniature helicopter,'' {\em Journal of Guidance
  Control and Dynamics}, vol.~27, no.~5, pp.~752--759, 2004.

\bibitem{frazzoli2005maneuver}
E.~Frazzoli, M.~A. Dahleh, and E.~Feron, ``Maneuver-based motion planning for
  nonlinear systems with symmetries,'' {\em Robotics, IEEE Transactions on},
  vol.~21, no.~6, pp.~1077--1091, 2005.

\bibitem{mettler2002rotorcraft}
B.~Mettler, M.~Valenti, T.~Schouwenaars, E.~Frazzoli, and E.~Feron,
  ``Rotorcraft motion planning for agile maneuvering,'' in {\em Proceedings of
  the 58th Forum of the American Helicopter Society, Montreal, Canada},
  vol.~32, 2002.

\bibitem{gibson1958visually}
J.~J. Gibson, ``Visually controlled locomotion and visual orientation in
  animals,'' {\em British journal of psychology}, vol.~49, no.~3, pp.~182--194,
  1958.

\bibitem{simon1972theories}
H.~A. Simon, ``Theories of bounded rationality,'' {\em Decision and
  organization}, vol.~1, pp.~161--176, 1972.

\bibitem{gibson1979ecological}
J.~Gibson, ``The ecological approach to human perception,'' 1979.

\bibitem{gibson1977theory}
J.~J. Gibson, ``The theory of affordances,'' {\em Perceiving, acting, and
  knowing: Toward an ecological psychology}, pp.~67--82, 1977.

\bibitem{warren1984perceiving}
W.~H. Warren, ``Perceiving affordances: visual guidance of stair climbing.,''
  {\em Journal of experimental psychology: Human perception and performance},
  vol.~10, no.~5, p.~683, 1984.

\bibitem{wilson2013embodied}
A.~D. Wilson and S.~Golonka, ``Embodied cognition is not what you think it
  is,'' {\em Frontiers in psychology}, vol.~4, 2013.

\bibitem{li2015towards}
B.~Li, B.~Mettler, and T.~M. Kowalewski, ``Towards data-driven hierarchical
  surgical skill analysis,'' {\em arXiv preprint arXiv:1503.08866}, 2015.

\bibitem{verma2016investigating}
A.~Verma and B.~Mettler, ``Investigating human learning and decision-making in
  navigation of unknown environments,'' {\em IFAC-PapersOnLine}, vol.~49,
  no.~32, pp.~113--118, 2016.

\bibitem{tseng2016human}
K.-S. Tseng and B.~Mettler, ``Human planning and coordination in spatial search
  problems,'' {\em IFAC-PapersOnLine}, vol.~49, no.~32, pp.~222--227, 2016.

\bibitem{andersh2014modeling}
J.~Andersh, B.~Li, and B.~Mettler, ``Modeling visuo-motor control and guidance
  functions in remote-control operation,'' in {\em Intelligent Robots and
  Systems (IROS 2014), 2014 IEEE/RSJ International Conference on},
  pp.~4368--4374, IEEE, 2014.

\bibitem{verma2015investigation}
A.~Verma, A.~Feit, and B.~Mettler, ``Investigation of human first-person
  guidance strategy from gaze tracking data,'' in {\em Systems, Man, and
  Cybernetics}, IEEE, 2015.

\bibitem{polani2011informational}
D.~Polani, ``An informational perspective on how the embodiment can relieve
  cognitive burden,'' in {\em Artificial Life (ALIFE), 2011 IEEE Symposium on},
  pp.~78--85, IEEE, 2011.

\bibitem{drugowitsch2012cost}
J.~Drugowitsch, R.~Moreno-Bote, A.~K. Churchland, M.~N. Shadlen, and A.~Pouget,
  ``The cost of accumulating evidence in perceptual decision making,'' {\em The
  Journal of Neuroscience}, vol.~32, no.~11, pp.~3612--3628, 2012.

\bibitem{braun2010structure}
D.~A. Braun, C.~Mehring, and D.~M. Wolpert, ``Structure learning in action,''
  {\em Behavioural brain research}, vol.~206, no.~2, pp.~157--165, 2010.

\bibitem{dijk2011grounding}
S.~G. van Dijk and D.~Polani, ``Grounding subgoals in information
  transitions,'' in {\em 2011 IEEE Symposium on Adaptive Dynamic Programming
  and Reinforcement Learning (ADPRL)}, pp.~105--111, IEEE, 2011.

\bibitem{ferrari2003clustering}
G.~Ferrari-Trecate, M.~Muselli, D.~Liberati, and M.~Morari, ``A clustering
  technique for the identification of piecewise affine systems,'' {\em
  Automatica}, vol.~39, no.~2, pp.~205--217, 2003.

\bibitem{krishnan2018transition}
S.~Krishnan, A.~Garg, S.~Patil, C.~Lea, G.~Hager, P.~Abbeel, and K.~Goldberg,
  ``Transition state clustering: Unsupervised surgical trajectory segmentation
  for robot learning,'' in {\em Robotics Research}, pp.~91--110, Springer,
  2018.

\bibitem{sun2009mining}
L.~Sun, R.~Patel, J.~Liu, K.~Chen, T.~Wu, J.~Li, E.~Reiman, and J.~Ye, ``Mining
  brain region connectivity for alzheimer's disease study via sparse inverse
  covariance estimation,'' in {\em Proceedings of the 15th ACM SIGKDD
  international conference on Knowledge discovery and data mining},
  pp.~1335--1344, ACM, 2009.

\bibitem{feit2015experimental}
A.~Feit and B.~Mettler, ``Experimental framework for investigating first-person
  guidance and perception,'' in {\em Systmes, Man, and Cybernetics}, IEEE,
  2015.

\bibitem{mettler2017emergent}
B.~Mettler, A.~Verma, and A.~Feit, ``Emergent patterns in agent-environment
  interactions as functional units supporting agile spatial skills,'' {\em
  Annual Reviews in Control}, 2017.

\bibitem{mettler2010agile}
B.~Mettler, N.~Dadkhah, and Z.~Kong, ``Agile autonomous guidance using spatial
  value functions,'' {\em Control Engineering Practice}, vol.~18, no.~7,
  pp.~773--788, 2010.

\bibitem{sutton1999mdps}
R.~S. Sutton, D.~Precup, and S.~Singh, ``Between mdps and semi-mdps: A
  framework for temporal abstraction in reinforcement learning,'' {\em
  Artificial intelligence}, vol.~112, no.~1, pp.~181--211, 1999.

\bibitem{lee1998guiding}
D.~N. Lee, ``Guiding movement by coupling taus,'' {\em Ecological psychology},
  vol.~10, no.~3-4, pp.~221--250, 1998.

\bibitem{liu2009slep}
J.~Liu, S.~Ji, J.~Ye, {\em et~al.}, ``Slep: Sparse learning with efficient
  projections,'' {\em Arizona State University}, vol.~6, no.~491, p.~7, 2009.

\bibitem{hayhoe2005eye}
M.~Hayhoe and D.~Ballard, ``Eye movements in natural behavior,'' {\em Trends in
  cognitive sciences}, vol.~9, no.~4, pp.~188--194, 2005.

\bibitem{li2015classification}
B.~Li, B.~Mettler, and J.~Andersh, ``Classification of human gaze in spatial
  guidance and control,'' in {\em Systems, Man, and Cybernetics (SMC), 2015
  IEEE International Conference on}, pp.~1073--1080, IEEE, 2015.

\end{thebibliography}

\end{document}